\documentclass[letterpaper, 10 pt, conference]{template/ieeeconf}  

\IEEEoverridecommandlockouts                              
\usepackage[utf8]{inputenc}
\usepackage[T1]{fontenc}

\usepackage{svg}    
\usepackage{amsmath} 
\usepackage{amssymb}  

\usepackage{mathtools}
\usepackage{subdepth} 
\usepackage{bm, upgreek} 

\usepackage{booktabs}
\usepackage{url}
\usepackage[hidelinks]{hyperref}

\usepackage{array} 
\newcolumntype{P}[1]{>{\centering\arraybackslash}p{#1}}
\newcolumntype{M}[1]{>{\centering\arraybackslash}m{#1}}

\usepackage{multirow}

\usepackage{caption}     
\usepackage{graphicx}    
\usepackage{subcaption}  

\usepackage{placeins}
\FloatBarrier

\title{\LARGE \bf
InterKey: Cross-modal Intersection Keypoints \\for Global Localization on OpenStreetMap
}

\author{Nguyen Hoang Khoi Tran, Julie Stephany Berrio, Mao Shan, and Stewart Worrall
}

\begin{document}

\twocolumn[{%
 \renewcommand\twocolumn[1][]{#1}
 \maketitle

 \vspace{-10px}
 \begin{center}
        \centering
    \includegraphics[width=1\linewidth]{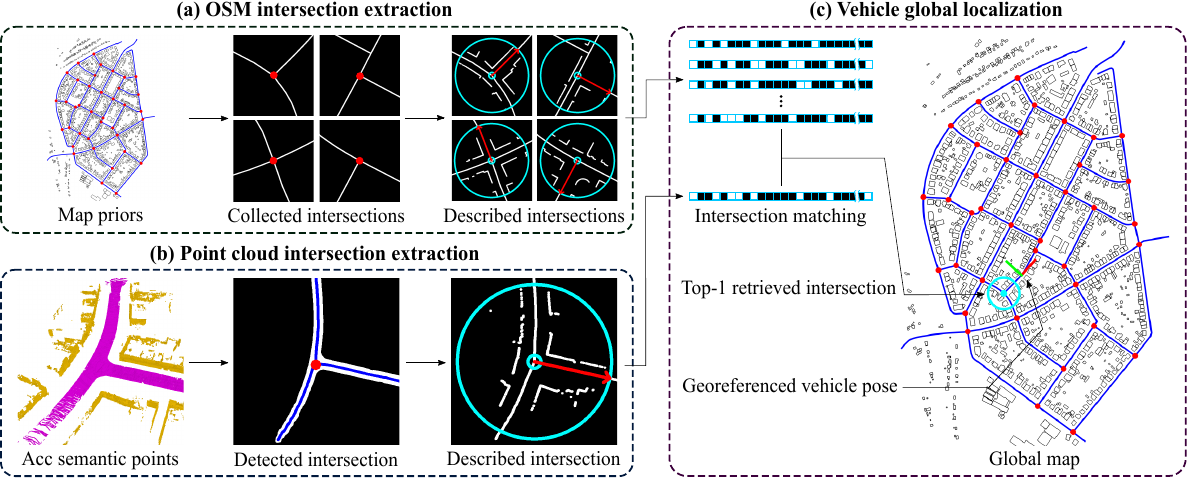}
    \captionof{figure}{Overall architecture of the proposed method. (a) We collect intersection nodes (red dots) from the map priors and encode the imprints of the surrounding roads (blue lines) and buildings (gray blocks) of each intersection into a binary descriptor. These descriptors together form the map intersection database. (b) We project the accumulated semantic point cloud of roads (magenta) and buildings (yellow) into top-view imprints, detect the road intersection (red dot), and generate its binary descriptor. (c) We retrieve the best-matched OSM intersection (cyan circle) by matching the descriptors and use it to georeference the vehicle pose (red and green axes).}
    \label{fig:ovr_pl}
\end{center}
 }]

\renewcommand{\thefootnote}{}  
\footnotetext{This work has been supported by the Vingroup Science and Technology Scholarship Program for Overseas Study for Master's and Doctoral Degrees (Corresponding author: Nguyen Hoang Khoi Tran). The authors are with the Australian Centre for Robotics (ACFR) at The University of Sydney (NSW, Australia). E-mails: \{\texttt{n.tran, j.berrio, m.shan, s.worrall}\}@acfr.usyd.edu.au.}

\thispagestyle{empty}
\pagestyle{empty}


\begin{abstract}

Reliable global localization is critical for autonomous vehicles, especially in environments where GNSS is degraded or unavailable, such as urban canyons and tunnels.
Although high-definition (HD) maps provide accurate priors, the cost of data collection, map construction, and maintenance limits scalability. OpenStreetMap (OSM) offers a free and globally available alternative, but its coarse abstraction poses challenges for matching with sensor data.
We propose InterKey, a cross-modal framework that leverages road intersections as distinctive landmarks for global localization. Our method constructs compact binary descriptors by jointly encoding road and building imprints from point clouds and OSM. To bridge modality gaps, we introduce discrepancy mitigation, orientation determination, and area-equalized sampling strategies, enabling robust cross-modal matching.
Experiments on the KITTI dataset demonstrate that InterKey achieves state-of-the-art accuracy, outperforming recent baselines by a large margin. The framework generalizes to sensors that can produce dense structural point clouds, offering a scalable and cost-effective solution for robust vehicle localization.

\end{abstract}


\section{Introduction}

Accurate global localization is essential for autonomous vehicles (AVs), supporting tasks such as initialization and recovery from localization failures, loop closure in simultaneous localization and mapping (SLAM), map-based path planning, and multi-vehicle coordination \cite{yin2024survey}. Conventionally, this capability relies on Global Navigation Satellite Systems (GNSS). However, GNSS signals are prone to degradation in urban canyons and forested areas due to multipath effects and signal blockage, resulting in unreliable positioning. This motivates the development of localization strategies that can operate robustly under GNSS-degraded conditions.

Recent research has demonstrated the effectiveness of high-definition (HD) maps in providing reliable priors for global localization \cite{elghazaly2023high}. Despite their utility, HD maps are costly to acquire, update, and maintain, which limits scalability. In contrast, OpenStreetMap (OSM) offers a compelling alternative: it 
is freely available, crowd-sourced, and globally maintained map with extensive coverage \cite{floros_openstreetslam_2013, osm_slam2023}. Although early studies have shown the feasibility of using OSM for global localization, a significant challenge remains in bridging the modality gap between coarse OSM representations and dense sensor-derived data collected by AVs \cite{yin2024survey}.

Among the structural elements available in OSM, road intersections are particularly important features for localization \cite{tran2025interloc}. Intersections exhibit distinctive geometric configurations that can be reliably extracted from LiDAR data, while simultaneously reducing the search space compared to general place recognition approaches. Despite this, existing cross-modal localization methods have not been explicitly designed to exploit intersections as primary features. Previous work has relied mainly on building contours as descriptors \cite{cho_openstreetmap-based_2022, li2024lidar}, which are susceptible to occlusion, repetitive patterns, or the absence of buildings, leading to ambiguous matches. Road geometry, although consistently available in OSM \cite{floros_openstreetslam_2013, habermann_road_2016}, remains underutilized in descriptor design.

In this work, we address the problem of global vehicle localization by leveraging cross-modal intersection recognition between point clouds and OSM. 
As a significant extension of our previous work \cite{tran2025interloc}, which focused on detecting road intersections in point clouds, \textbf{InterKey} tackles the complementary problem of cross-modal intersection matching between point clouds and OSM for global localization. 
Specifically, \textbf{InterKey} is a novel framework that uses road and building imprints around intersections to generate compact and distinctive descriptors. To bridge the modality gap, we introduce a discrepancy mitigation method along with orientation determination and shape encoding strategies, enabling robust descriptor matching across point clouds and OSM data. By localizing intersections in point clouds and aligning them with OSM priors, our framework estimates the local-to-global coordinate transformation and projects the vehicle pose into the global reference frame. Fig. \ref{fig:ovr_pl} presents the overall architecture of the framework.

The main contributions of this paper are threefold.
\begin{itemize}
   \item  We present the first framework that explicitly leverages cross-modal intersection matching for point-cloud-to-OSM global localization, demonstrating its effectiveness in GNSS-denied or degraded environments.
    \item We propose a novel intersection descriptor that integrates road and building information, accompanied by a modality discrepancy mitigation scheme, an orientation determination strategy, and an area-equalized sampling approach to improve robustness and efficiency.
    \item We achieve state-of-the-art performance compared to existing OSM-based localization approaches and feature description methods, as validated through experiments on real-world datasets.
\end{itemize}

By using intersections as global landmarks and leveraging the scalability of OSM, our method provides a practical and cost-effective solution for AV localization at scale.

\section{Related Work}

Point-cloud-based global localization has gained prominence as a solution for GNSS-denied or degraded environments. 
This section reviews state-of-the-art methods along two main paradigms: point-cloud-to-point-cloud and point-cloud-to-OSM approaches.

\subsection{Point-Cloud-to-Point-Cloud Approaches}

Several approaches focus on directly matching point clouds without using external maps. Scan Context \cite{kim_scan_2018} and its extension Scan Context++ \cite{kim_scan_2022} introduced bird's-eye view (BEV) descriptors for large-scale place recognition, showing robustness against viewpoint changes. Deep learning methods such as PointNetVLAD \cite{uy2018pointnetvlad}, Locus \cite{vidanapathirana2021locus}, and LoGG3D-Net \cite{vidanapathirana2022logg3d} further advanced this area of research by learning global descriptors from point clouds. More recently, Wild-Places \cite{knights2022wild} and A New Horizon \cite{shi2024new} explored recognition in unstructured or clustered environments. 
Furthermore, Semantic Scan Context \cite{li2021ssc} incorporates semantic labels into the Scan Context framework, demonstrating the value of semantic understanding in place recognition tasks.

Although these methods are effective for loop closure and place recognition, they rely on prior point cloud maps and therefore lack global reference information, limiting their applicability for GNSS-denied global localization.

\subsection{Point-Cloud-to-OSM Approaches}

Another line of research involves developing cross-modal descriptors to match 3D point clouds with 2D map features. 
The key challenge lies in the modality gap: sparse 3D point clouds versus coarse 2D abstractions in OSM. 
OSM offers lightweight but globally referenced elements such as roads and buildings. Recent work leverages these abstractions for global pose estimation, enabling open-world localization and drift correction. 
Early contributions such as OpenStreetSLAM \cite{floros_openstreetslam_2013} and subsequent work on localization with OSM data \cite{ruchti2015localization} aligned odometry or road points to road graphs. To improve distinctiveness, \cite{yan2019global} proposed 4-bit intersection/building descriptors, while the SLAM Graph Matching Pose \cite{he2019graph} exploited the road topology.

Recent methods address point-cloud-to-OSM matching by projecting both sources of information into a common representation, often via polar histograms.
OSM Context \cite{cho_openstreetmap-based_2022} matches global poses by comparing the angular distance descriptors of the building contours in OSM with the corresponding ranges of the LiDAR scans, removing the need for a prior LiDAR map.
Recent work improves hand-crafted descriptors by enhancing the robustness to map errors. For example, \cite{li2024lidar} propose a boundary relative orientation descriptor that encodes boundary trends, offering rotational consistency and scale invariance against map or GNSS errors.

Beyond buildings, researchers have also exploited the geometry of OSM roads. In \cite{elhousni2022lidar}, the authors used a particle filter restricted by OSM road lanes and junctions to globally localize LiDAR scans, treating the map as a set of geometric constraints to guide the hypothesis. 
Similarly, \cite{ballardini2021vehicle} matched LiDAR-derived 3D point clouds of building facades to 3D building models from OSM, achieving meter-level precision by aligning structural contours.

Beyond hand-crafted features, learning-based methods aim to bridge the LiDAR–map gap by training deep networks to learn shared latent representations for cross-modal matching \cite{yin2024survey}. For example, \cite{lee2024autonomous} uses ResNet-18 with NetVLAD, and OPAL \cite{kang2025opal} learns joint LiDAR–OSM descriptors using a Siamese network with visibility masking and adaptive radial fusion, achieving state-of-the-art cross-modal localization. OpenLiDARMap \cite{kulmer2025openlidarmap} aligned LiDAR scans with OSM footprints and surface models to eliminate drift for point cloud mapping without GNSS. 

Unlike prior methods, our framework takes advantage of matching robust intersection features to address point-cloud-to-OSM global localization. We combine road and building information into a unified descriptor to enhance distinctiveness and introduce a novel area-equalized sampling pattern to mitigate range bias in descriptor construction. Together, these innovations help bridge the modality gap between sensor-derived point clouds and OSM priors, enabling robust cross-modal intersection matching and global localization.

\section{Methodology}


This section presents our proposed method for vehicle global localization using point-cloud-to-OSM intersection matching.
The method consists of three stages: (1) extracting an intersection descriptor database from OSM data, (2) extracting the query intersection descriptor from point cloud data, and (3) localizing the vehicle in the global reference system.
To bridge the modality gap between OSM and point cloud data, we preprocess the inputs in steps (1) and (2) into a unified form, where each intersection is represented by its position and the imprints of surrounding roads and buildings. 
From this unified representation, we apply the same cross-modal intersection description algorithm in the two steps to generate the descriptor database and the query descriptor.
In step (3), we match these descriptors to identify the most similar map intersection and infer the vehicle's global pose.
Fig. \ref{fig:ovr_pl} visualizes the overall architecture of the method. 

\subsection{OSM Data Preprocessing} \label {sec:osm_proc}

This module processes the raw OSM data to produce the map intersection positions together with the surrounding road and building imprints. 
Suppose we have a rough estimate of the vehicle’s georeferenced location, obtained from sources such as GNSS, cellular networks, or other infrastructure. We load the map data of the local area from OSM and extract the map priors, including the road graph and building polygons. We perform this extraction using off-the-shelf tools detailed in Section \ref{sec:exp_eval_dat}. The prior road graph consists of edges representing drivable road segments and nodes representing their junctions. The building polygons include vertex coordinates that approximate the building contours.

\subsubsection{Intersection position collection}
We extract nodes with at least three connecting edges and define them as intersection nodes. Let $(Y_l)$ be the coordinate frame of the $l$-th map intersection node. We collect the positions of map intersection nodes in the global reference frame $(G)$ as $\{{_G}{\mathbf{p}}_{Y_l}\}$, where $\mathbf{p} \in \mathbb{R}^2$ denotes a position on ground plane.

\subsubsection{Road imprint generation}
We find the local road subgraph at each intersection. Starting from the intersection node, we trace the prior road graph using depth-first search until reaching a node that lies outside the image range or has no more connecting edges. 
We then plot the road segments in each subgraph as one-pixel-wide objects on a binary image centered at the intersection node. We obtain the set of road imprints as $\{ \mathbf{I}^{\mathrm{mr}}_l \in {\mathbb{B}}^{M \times M} \}$, where $\mathbb{B}=\{0,1\}$ is the binary set, and $M$ is determined by the image physical size $S$ and resolution $r$.

\subsubsection{Building imprint generation}
We search for local buildings around each intersection by selecting the subset of building polygons that have at least one vertex within the image range.
For each intersection, we then draw one-pixel-wide lines between the vertices of these polygons on a binary image centered at the intersection node. We obtain the set of building imprints as $\{ \mathbf{I}^{\mathrm{mb}}_l \in {\mathbb{B}}^{M \times M} \}$. Each image has the same size and resolution as the corresponding road image.

\subsection{Point Cloud Data Preprocessing} \label{sec:pcl_proc}
This module handles the raw point cloud data to generate the observed intersection position and the imprints of surrounding roads and buildings. 
We assume the availability of vehicle odometry and semantic segmentation for roads and buildings. These data are typically provided by vehicle navigation and perception packages, as mentioned in Section \ref{sec:exp_eval_dat}. We create the accumulated road-and-building point cloud by applying the semantic mask and concatenating a window of $K$ point clouds called keyframes. A keyframe is selected when its position or heading differs from the previous keyframe by at least $D^\mathrm{p}$ or $D^\mathrm{h}$, respectively.

\subsubsection{Road imprint generation}
We project the accumulated road points onto a top-view binary image. This image is centered on the vehicle coordinate frame of the middle keyframe among the $K$ keyframes, denoted as $(L_c)$, where $c$ is the point cloud index. We infer the road surface using morphological closing and opening, then skeletonize it into a one-pixel-wide centerline. The resulting image forms the point cloud road imprint, denoted as $\mathbf{I}^{\mathrm{pr}}_c \in {\mathbb{B}}^{M \times M}$.

\subsubsection{Intersection point detection}
We detect intersection points on the road imprint using the Harris corner detector. For simplicity, we retain only the detection point closest to the vehicle position. Let $(I_j)$ denote the coordinate frame of the $j$-th detected intersection. We obtain the observed intersection position in the vehicle-centric reference frame as $_{L_c}\mathbf{p}_{I_j} \in \mathbb{R}^2$.

\subsubsection{Building imprint generation}
We project the accumulated building points onto a top-view binary image centered at $(L_c)$. This forms the point cloud building imprint, $\mathbf{I}^{\mathrm{pb}}_c \in {\mathbb{B}}^{M \times M}$. Both the point cloud road and building imprints share the same size and resolution as their map counterparts.
 
\subsection{Cross-modal Intersection Description} \label{sec:its_desc}

This module describes each intersection as a unique binary string to facilitate matching, using the preprocessed intersection data from either the map or the point cloud. During the description, the module also refines the intersection position and estimates its orientation, which are useful for vehicle global localization in the next stage. 

\begin{figure} [t]
    \centering
    \begin{subfigure}{0.325\columnwidth}
        \includegraphics[width=\linewidth]{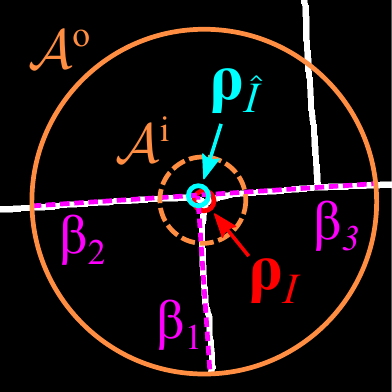}
        \caption{\centering Discrepancy mitigation}
        \label{fig:its_desc_a}
    \end{subfigure}
    \begin{subfigure}{0.325\columnwidth}
        \includegraphics[width=\linewidth]{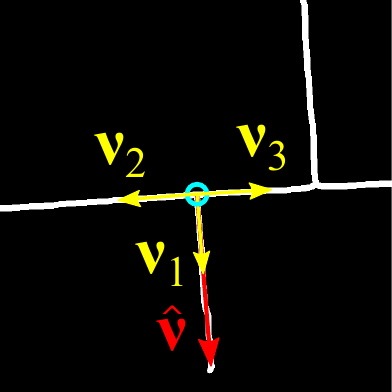}
        \caption{\centering Orientation determination}
        \label{fig:its_desc_b}
    \end{subfigure}
    \begin{subfigure}{0.325\columnwidth}
        \includegraphics[width=\linewidth]{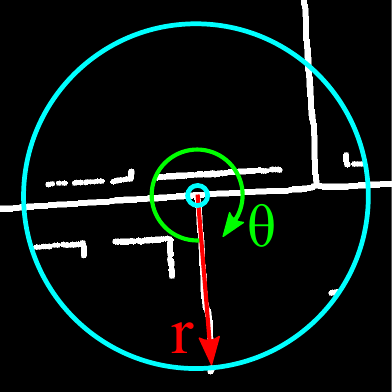}
        \caption{\centering Shape\\encoding}
        \label{fig:its_desc_c}
    \end{subfigure}
    \vskip0.2em 
    \begin{subfigure}{1\columnwidth}
        \includegraphics[width=\linewidth]{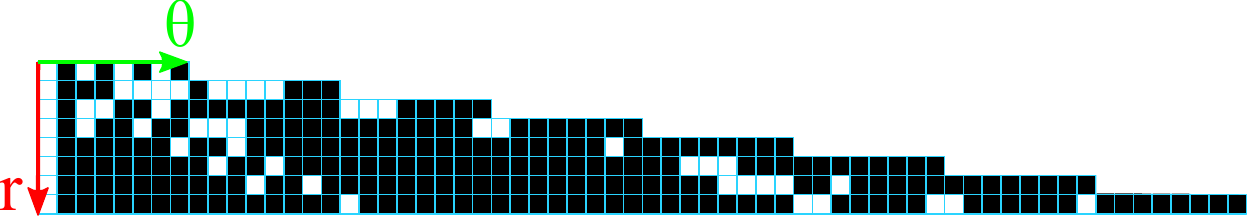}
        \caption{Generated descriptor}
        \label{fig:its_desc_d}
    \end{subfigure}
    \caption{Intersection description process. (a) We adjust the original intersection point (red) to the refined intersection point (cyan) and correct the road imprint in the discrepant region (orange dashed) based on the approximate branch lines (magenta). (b) We select the characteristic orientation (red) from the branch orientations of the intersection (yellow). (c) We apply the sampling pattern to the combined road and building imprints. The pattern (large cyan circle) is centered at the refined intersection point (small cyan circle) and aligned with the characteristic orientation (red arrow). (d) We assemble the sampled cells in the order of ring (red arrow) and angle (green arrow) to generate the descriptor.}
    \label{fig:its_desc}
    \vspace{-5mm}
\end{figure}

Although preprocessed into a unified representation, the intersection data from OSM and point clouds still contain certain discrepancies in values. This reduces the reliability of describing the same intersection across the two modalities, leading to incorrect matches. We first mitigate cross-modal discrepancies in the intersection data, including the intersection position and the imprints of roads and buildings.
To achieve rotation invariance in the description, we determine a characteristic orientation for each intersection to steer the sampling pattern. We compute this orientation using only the road imprint, which is less affected by occlusion and segmentation errors than the building imprint in practice.
Finally, we sample the descriptor string that represents the local shape of the intersection using the combined imprint of roads and buildings.
Fig. \ref{fig:its_desc} summarizes the main steps of this module.

\subsubsection{Cross-modal discrepancy mitigation} \label{sec:diff_miti}
%

Let $\bm{\uprho} \in \mathbb{R}^2$ denote a point in the image coordinate system. We define two regions in the preprocessed road imprint around the preprocessed intersection point $\bm{\uprho}_I$. The consistent region $\mathcal{A}^\mathrm{o}$ is an annulus with inner radius $R^\mathrm{i}$ and outer radius $R^\mathrm{o}$. The discrepant region $\mathcal{A}^\mathrm{i}$ is a disk located inside and adjacent to the consistent region.
We segment the intersection branches by finding continuous objects in the consistent region. Each branch starts from the inner boundary of the consistent region and ends before meeting another intersection point. We obtain the set of branch starting points $\{\bm{\uprho}^\mathrm{s}_b\}$ and the set of branch centroids $\{\bm{\uprho}^\mathrm{c}_b\}$, where $b \in [1,B]$ is the branch index and $B$ is the number of branches. For each branch, we define the approximate branch line $\upbeta_b$ that passes through its starting point and centroid. 

We compute the refined intersection point $\bm{\uprho}_{\hat{I}}$ as the point within the discrepant region that minimizes the sum of squared perpendicular distances to the approximate branch lines. 
We create the refined road imprint $\hat{\mathbf{I}}^\mathrm{r}$ by copying the consistent region from the preprocessed road imprint $\mathbf{I}^\mathrm{r}$ and, in the discrepant region, drawing lines from the refined intersection point to the branch starting points. 
We generate the refined building imprint $\hat{\mathbf{I}}^\mathrm{b}$ by ray casting the preprocessed building imprint $\mathbf{I}^\mathrm{b}$ from the refined intersection point. We summarize these adjustments as follows:
\begin{align}
    \bm{\uprho}_{\hat{I}} &= \underset{ \bm{\uprho} \in \mathcal{A}^\mathrm{i} }{ \arg\!\min} \! \left( \sum_{b=1}^{B}{\delta \left( \bm{\uprho}, \upbeta_b \right)^2} \right) \\
    \hat{\mathbf{I}}^{\mathrm{r}} &= \sigma ( \mathbf{I}^{\mathrm{r}}, \bm{\uprho}_{\hat{I}}, \{ \bm\uprho^\mathrm{s}_b \} ) \\
    \hat{\mathbf{I}}^{\mathrm{b}} &= \phi ( \mathbf{I}^{\mathrm{b}}, \bm{\uprho}_{\hat{I}} )
    \label{eqn:disc_miti}
\end{align}
where $\delta(\cdot)$ is the perpendicular distance function from a point to a line, $\sigma(\cdot)$ is the road imprint refinement function, and $\phi(\cdot)$ is the ray-casting function.
Fig. \ref{fig:its_desc_a} illustrates the adjustment process.

\subsubsection{Orientation determination} \label{sec:desc_orient}

We define the orientation of each branch as the normalized vector from the refined intersection point to its centroid. The set of branch orientation vectors is $\mathcal{V} = \{\bm{\upnu}_b\}$. We compute the summary orientation vector as $\bm{\upnu}^\mathrm{s}=\Sigma^B_{b=1}\bm{\upnu}_b$.
We use the summary vector to select the characteristic orientation from the branch orientations rather than define it directly. This is because opposite branch orientations cancel each other out when summed, making the resulting orientation sensitive to noise.

When the intersection is asymmetric, we define the characteristic orientation $\hat{\bm{\upnu}}$ as the branch orientation with the smallest angular distance to the summary vector as follows:
\begin{equation}
    \hat{\bm{\upnu}} = \underset{{\bm{\upnu}}_b \in \mathcal{V}}{\arg\!\min} \: \angle({\bm{\upnu}}_b,\bm{\upnu}^\mathrm{s}), \quad \text{when } \Vert \mathbf{\bm{\upnu}}^\mathrm{s} \rVert > \tau^\mathrm{s} 
    \label{eqn:char_orient}
\end{equation}
where $\angle(.)$ is the angle between two vectors, and $\tau^\mathrm{s}$ is the symmetry threshold considered on the magnitude of the summary vector.
In the case of symmetric intersections, for point cloud intersection extraction, we select the branch orientation with the smallest angle in the image coordinate system as the characteristic orientation. For OSM intersection extraction, we treat all branch orientations as characteristic orientations, generating multiple descriptors for each intersection in the database. This produces a single query and multiple candidates, allowing the matching process to automatically choose the orientation with the highest similarity.
Fig. \ref{fig:its_desc_b} visualizes the determined characteristic orientation.

\subsubsection{Shape encoding} \label{sec:desc_shape}

We define a circular sampling pattern with radius $R^\mathrm{o}$, divided into $N^\mathrm{r}$ equally spaced rings. Each ring is divided into $N^\mathrm{b}i$ cells of equal area, where $N^\mathrm{b}$ is the base number of cells, and $i \in [1,N^\mathrm{r}]$ is the index of the ring from inside to outside.
The cell area in the $i$-th ring is given by:
\begin{equation}
    a_i = \kappa \left( 2 - \frac{1}{i} \right), \quad \kappa = \frac{\pi {R^\mathrm{o}}^2}{N^\mathrm{b} {N^\mathrm{r}}^2} \;\; \text{(constant)}
    \label {eqn:pat_area}
\end{equation}
We create the combined imprint $\hat{\mathbf{I}}^\mathrm{c}$ as the logical OR between the two refined imprints of roads and buildings.
We apply the pattern on the combined imprint and assign each cell a binary value. The pattern is centered at the refined intersection point and aligned with the characteristic orientation. 
We concatenate the sampled values from the inner to outer rings, forming the binary descriptor string as:
\begin{equation}
     \mathbf{d}= \psi(\hat{\mathbf{I}}^{\mathrm{c}},  \bm{\uprho}_{\hat{I}}, \hat{\bm{\upnu}}) \in \mathbb{B}^Q
    \label {eqn:des_string}
\end{equation}
where $\psi(\cdot)$ denotes the shape encoding function, and $Q$ is determined by $N^\mathrm{b}$ and $N^\mathrm{r}$.
Fig. \ref{fig:its_desc_c} visualizes the sampling pattern applied on the combined imprint. Fig. \ref{fig:its_desc_d} shows the sampled descriptor unfolded along the ring and angle axes.
 
\begin{figure} [t]
    \centering
    \begin{subfigure}{0.1\columnwidth}
    \end{subfigure}
    \begin{subfigure}{0.39\columnwidth}
        \centering
        \includegraphics[width=1\linewidth]{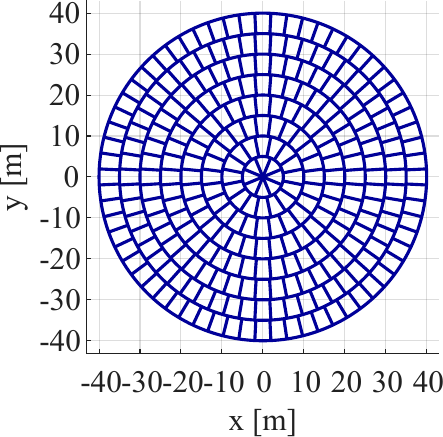}
        \caption{InterKey (ours)}
        \label{fig:sam_pat_ikey}
    \end{subfigure}
    \begin{subfigure}{0.39\columnwidth}
        \centering
        \includegraphics[width=1\linewidth]{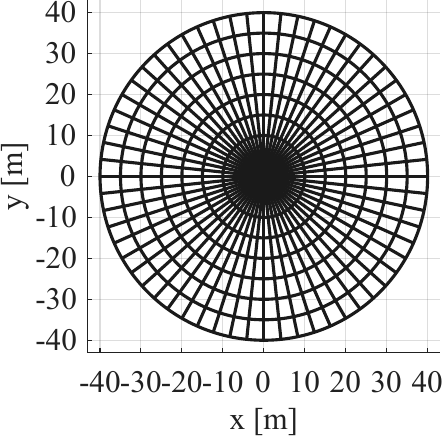}
        \caption{Scan Context}
        \label{fig:sam_pat_scon}
    \end{subfigure}
    \begin{subfigure}{0.1\columnwidth}
    \end{subfigure}
    \vskip1em 
    \begin{subfigure}{1\columnwidth}
        \centering
        \includegraphics[width=0.63\linewidth]{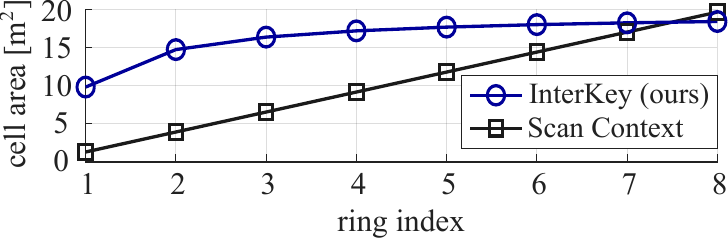}
        \caption{Cell area comparison}
        \label{fig:sam_pat_area}
    \end{subfigure}
    \caption{Sampling pattern of the proposed method (a) compared with Scan Context (b). The proposed pattern yields a more equalized distribution of cell areas across the rings than Scan Context (c).}
    \label{fig:sam_pat}
    \vspace{-5mm}
\end{figure}

Fig. \ref{fig:sam_pat} shows a comparison between the sampling pattern of the proposed method and that of the Scan Context \cite{kim_scan_2022}. The key difference lies in the distribution of cell areas across the rings. In Scan Context, the cell area increases linearly with ring index, whereas in our method it quickly saturates to a constant value, as mathematically shown by Eq. (\ref{eqn:pat_area}). This enables our method to capture the surrounding shapes with the same level of detail, whether near or far from the center. Such a property is particularly suitable for a point cloud accumulation approach like ours, where the spatial distribution of points becomes more uniform. In contrast, Scan Context is better suited for single point cloud scenarios, where the point density decreases from the center outward.

\subsection{Vehicle Global Localization}

This module initializes the vehicle pose in the global reference frame using point-cloud-to-map intersection matching. 
We query the descriptor database to identify the map intersection that best matches the observed intersection.
Using the best-matched map intersection as a landmark, we convert the vehicle pose from the local to the global reference frame.

\subsubsection{Intersection Matching} \label{sec:its_mat}

We retrieve the database descriptor that has the highest similarity to the query descriptor.
Since descriptor strings are binary, we define their similarity by the Hamming distance and organize the database into a hash table for efficient matching.

\subsubsection{Vehicle Pose Georeferencing}
We use the best match to georeference the vehicle pose when its Hamming distance falls within the decision threshold $\tau^\mathrm{h}$.
This match establishes the equivalence between the observed intersection and the map intersection, $(\hat{I}_j)=(\hat{Y}_l)$ .
We define the pose of an intersection as its refined position and characteristic orientation. 
From the description process in Section \ref{sec:its_desc}, we obtain the map intersection pose in the global reference frame, $\mathbf{T}_{G\hat{Y}_l} \in SE(2)$,
and the observed intersection pose in the middle-keyframe vehicle frame, $\mathbf{T}_{L_c\hat{I}_j} \in SE(2)$.
From odometry input, we obtain the pose of the current vehicle frame $(L_k)$ relative to the middle-keyframe vehicle frame as $\mathbf{T}_{{L_c}{L_k}} \in SE(2)$.
By combining these three transformations, we derive the current vehicle pose in the global reference frame as:
\begin{equation} 
        {\mathbf{T}}_{GL_k} = \mathbf{T}_{G{\hat{Y}_l}}  {\mathbf{T}^{-1}_{L_{c}\hat{I}_j}} \mathbf{T}_{{L_c}{L_k}}
    \label{eqn:dtc_its}
\end{equation}

\begin{figure*}[t!]
    \centering
    \begin{subfigure}{0.245\textwidth}
        \includegraphics[width=\linewidth]{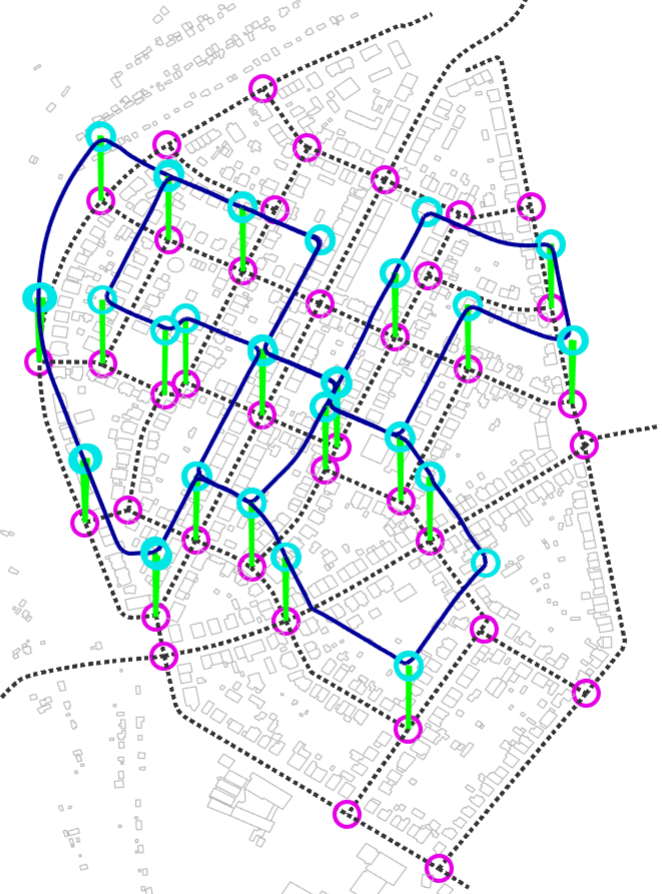}
        \caption{InterKey (ours)}
    \end{subfigure}
    \begin{subfigure}{0.245\textwidth}
        \includegraphics[width=\linewidth]{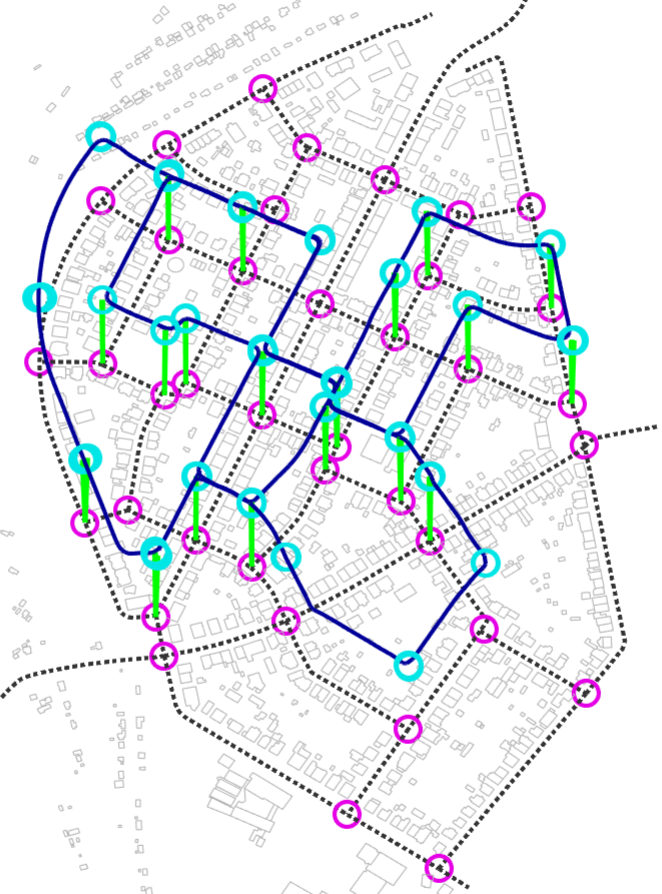}
        \caption{Scan Context}
    \end{subfigure}
    \begin{subfigure}{0.245\textwidth}
        \includegraphics[width=\linewidth]{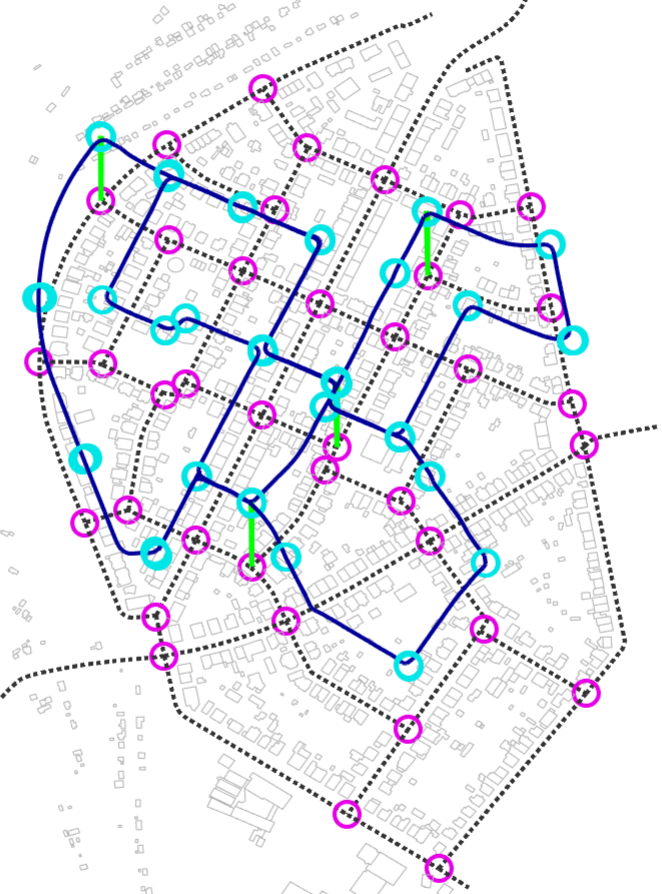}
        \caption{BRISK}
    \end{subfigure}
    \begin{subfigure}{0.245\textwidth}
        \begin{subfigure}{\linewidth}
            \includegraphics[width=\linewidth]{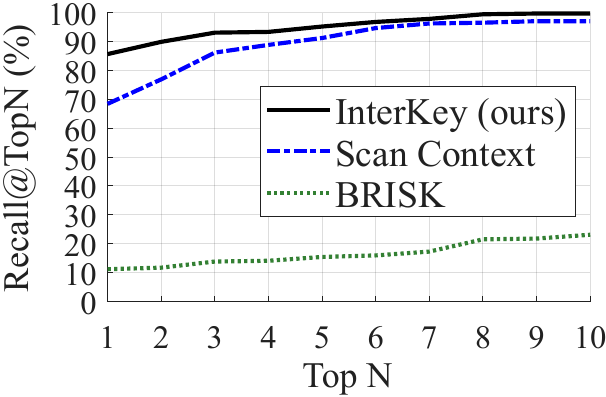}
        \end{subfigure}
        \vskip0.2em 
        \begin{subfigure}{\linewidth}
            \includegraphics[width=\linewidth]{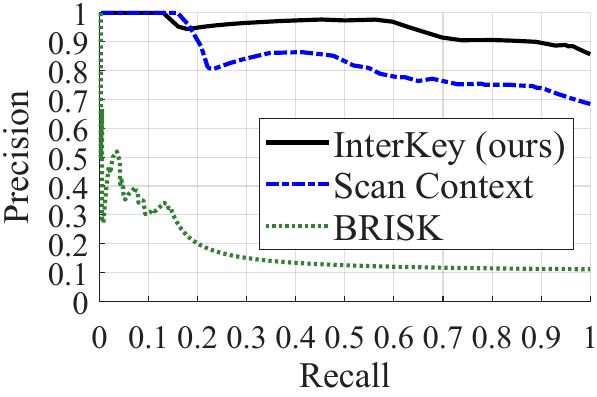}
        \end{subfigure}
        \caption{Metric comparison}
    \end{subfigure}
    \caption{Intersection matching performance of our method (a) compared to the two baselines (b), (c) on the KITTI 00 sequence. Lime green lines indicate true matches between the observed intersections (cyan circles) and the map intersections (magenta circles) with Top-1 retrieval. (d) Our method outperforms the baselines in both Recall@Top$N$ and Precision-Recall curve metrics. The solid, dark blue line represents the vehicle's trajectory. For clarity, this trajectory is rendered slightly above the OSM road map, which makes the lime green match lines easier to distinguish from the map background.}
    \label{fig:mat_eval_viz}
    \vspace{-5mm}
\end{figure*}


\section{Experimental Evaluation} \label{sec:exp_eval}

We conducted experiments to evaluate the performance of our method in terms of intersection matching and global localization. For intersection matching, we compare it with two well-known feature descriptor methods, BRISK \cite{leutenegger2011brisk} and Scan Context \cite{kim_scan_2022}. For vehicle global localization, we compare the method with two state-of-the-art methods, OSM Context \cite{cho_openstreetmap-based_2022} and Boundary Directional Feature (BDF) \cite {li2024lidar}. 

\begin{table}[t]
    \centering
    \caption{Experimental dataset information}
    \resizebox{\columnwidth}{!} {
        \begin{tabular}{l@{\hspace{3pt}} | c@{\hspace{3pt}} c@{\hspace{3pt}} c@{\hspace{3pt}} c@{\hspace{3pt}} c@{\hspace{3pt}} c@{\hspace{3pt}} c@{\hspace{3pt}} c@{\hspace{3pt}}} 
            \toprule
            Sequence & 00 & 02 & 05 & 06 & 07 & 08 & 09 & 10 \\
            \midrule
            Traveled distance [km] & 3.72 & 5.06 & 2.21 & 1.24 & 0.69 & 3.22 & 1.70 & 0.92 \\
            \#LiDAR point clouds & 4541 & 4661 & 2761 & 1101 & 1101 & 4071 & 1591 & 1201 \\
            \#OSM intersection nodes & 39 & 45 & 37 & 37 & 39 & 28 & 42 & 42 \\
            \bottomrule
        \end{tabular}
    }
    \label{tab:data_seqs}
\end{table}

\begin{table}[t]
    \centering
    \caption{Experimental parameters of the proposed algorithm}
        \begin{tabular}{c@{\hspace{3pt}} | l@{\hspace{3pt}} | r@{\hspace{2pt}} c@{\hspace{2pt}} l@{\hspace{2pt}}} 
            \toprule
            Module & Parameter & \multicolumn{3}{c}{Value} \\
            \midrule
            & Top-view image size & $S$ & $=$ & $120$ m \\
            Pre-& Top-view image resolution & $r$ & $=$ & $0.16$ m/px \\
            processing & Number of accumulated keyframes & $K$ & $=$ & $21$ \\
            & Keyframe position threshold & $D^\mathrm{p}$ & $=$ & $5$ m\\
            & Keyframe heading threshold & $D^\mathrm{h}$ & $=$ & $10^\circ$ \\
            \midrule
            & Descriptor outer radius & $R^\mathrm{o}$ & $=$ & $40$ m \\
            & Discrepant region radius & $R^\mathrm{i}$ & $=$ & $10$ m \\
            Description & Intersection symmetry threshold & $\tau^\mathrm{s}$ & $=$ & $0.1$ \\
            & Number of pattern rings & $N^\mathrm{r}$ & $=$ & 8 \\
            & Number of pattern base cells & $N^\mathrm{b}$ & $=$ & 8 \\
            \midrule
            Localization & Matching decision threshold & $\tau^\mathrm{h}$ & $=$ & $50$ \\
            \bottomrule
        \end{tabular}
    \label{tab:param_set}
    \vspace{-5mm}
\end{table}

\subsection{Dataset and Experimental Settings} \label{sec:exp_eval_dat}

We used the KITTI public dataset \cite{geiger2013vision} for experiments. Building on previous studies \cite{cho_openstreetmap-based_2022, li2024lidar}, we selected eight data sequences that contain real-world driving scenarios in residential, urban, and rural areas. In the dataset, we utilized roof-mounted LiDAR scans at a rate of 10 Hz and GNSS/IMU measurements at a rate of 100 Hz. We obtained the predicted semantic labels from the raw LiDAR scans using the RangeNet++ algorithm \cite{milioto2019rangenet}. We used the LIO-SAM framework \cite{shan2020lio} without loop closure to estimate vehicle odometry from LiDAR and inertial data. The RTK-corrected GNSS/IMU integration provided the ground-truth vehicle pose in the global coordinate system.

We collected map data from OSM. For each driving sequence, we exported a rectangular map area that completely covers the ground-truth trajectory of the vehicle. From the raw map file, we extracted the road graph using the \texttt{osm2pgrouting} tool. To obtain drivable road segments, we looked for `highway' objects of type `residential', `service', or `tertiary' with a non-empty name field.
Additionally, from the raw map file, we utilized the \texttt{osmium} tool to extract the `building' objects along with their corresponding contour coordinates. 
We summarize the experimental data in Table \ref{tab:data_seqs}. 
We set the method parameters for the experiments as listed in Table \ref{tab:param_set}.

\subsection{Intersection Matching Performance} \label{sec:exp_eval_mat}

\subsubsection{Evaluation Metrics}

To verify the correctness of an intersection match, we compute the Euclidean distance between each query intersection and its retrieved counterpart. We transform the position of the query intersection from the vehicle-centric reference frame to the global reference frame, which is also used for the retrieved intersection. We define the Euclidean distance for matching verification as follows.
\begin{equation}
    \Delta^{\mathrm{e}}_{jl} = \vert| \overline{\mathbf{T}}_{G{L_c}} {_{L_c}}\mathbf{p}_{\hat{I}_j} - {_{G}}\mathbf{p}_{\hat{Y}_l} \vert|
\end{equation}
where $\overline{\mathbf{T}}_{G{L_c}}$ is the ground-truth vehicle pose provided by the test dataset. If this distance is smaller than $5$ m, the match is considered true.

We used standard metrics of Recall@Top$N$ and precision-recall curve to evaluate the method, following  \cite{lee2024autonomous, cho_openstreetmap-based_2022, kim_scan_2022}.
Suppose that each query returns $N$ best-matched candidates. We consider a correct retrieval if at least one of the $N$ candidates is a true match to the query. We define Recall@Top$N$ as the ratio of correct retrievals to the total number of queries.
In addition, the precision-recall curve is calculated for the Top-1 candidate under different decision thresholds of the descriptor distance. 

\subsubsection{Baselines}

To the best of our knowledge, there is currently no equivalent work in our specific research scope, point-cloud-to-OSM intersection matching. Therefore, we adopt two popular feature description methods from related fields for comparison. One is Scan Context \cite{kim_scan_2022}, a structural descriptor for point-cloud-to-point-cloud recognition. The other is BRISK \cite{leutenegger2011brisk}, a keypoint descriptor for image-to-image matching. 
For fair comparison, we provided all methods with the same input data: the positions of the intersection points to be described and their corresponding combined imprints of roads and buildings. These inputs were given by the processes presented in Section \ref{sec:its_desc}.
We implemented the two comparative methods as follows:

\textbf{Scan Context.} We modified the Scan Context cell data type to binary to accommodate the 2D image input, following \cite{cho_openstreetmap-based_2022}. For distance calculation between two feature vectors, we retained the original column-shift search but replaced the distance metric with Hamming distance to better suit the binary data type. We configured the Scan Context sampling pattern to be similar to our method, with a range of $40$ m, $8$ rings, and $60$ sectors. Fig. \ref{fig:sam_pat_scon} visualizes this configuration.

\textbf{BRISK.} We scaled the sampling pattern of BRISK so that its radius reaches $40$ m to be equivalent to our method. We kept the rotation estimation method as the original. 

\subsubsection{Results}

\begin{table}[t]
    \centering
    \caption{Recall@Top$N$ (\%) of intersection matching averaged over KITTI dataset and the feature vector size (bytes) of methods.}
    \begin{tabular}{c | c c c | c} 
        \toprule
        \multirow{2}{*}{Description method} & \multicolumn{3}{c|}{Recall@Top$N$} & \multirow{2}{*}{Feature size}\\
        \cmidrule{2-4}
        & Top 1  & Top 5 & Top 10 & \\
        \midrule
        BRISK \cite{leutenegger2011brisk} & 16.73 & 31.38 & 42.00 & 64 \\
        Scan Context \cite{kim_scan_2022} & 42.85 & 69.16 & 74.72 & 60 \\
        \midrule
        InterKey (ours) & \textbf{67.81} & \textbf{86.09} & \textbf{90.70} & \textbf{36} \\
        \bottomrule
    \end{tabular}
    \label{tab:mat_acc}
    \vspace{-5mm}
\end{table}

Figs. \ref{fig:mat_eval_viz}a to \ref{fig:mat_eval_viz}c present a qualitative comparison of intersection matching performance across the methods. Among the baselines, BRISK achieves only a few true matches relative to the number of detected intersections, whereas Scan Context achieves a larger proportion. The proposed method obtains the highest number of true matches and outperforms both baselines.
Fig. \ref{fig:mat_eval_viz}d shows a quantitative comparison of the methods in terms of Recall@Top$N$ and the precision-recall curve. The proposed method achieves a higher rate of correct retrievals than the baselines in all top candidate counts from 1 to 10. Across different decision thresholds, it maintains higher precision than the baselines at most recall levels.
Table \ref{tab:mat_acc} shows that the proposed method achieves a higher performance despite having a smaller descriptor string size than the baselines.
These results indicate that the proposed method is more accurate and efficient than the baselines in point-cloud-to-OSM intersection matching.

\begin{figure*} [t]
    \centering
    \begin{subfigure}[t]{0.245\linewidth}
        \centering
        \includegraphics[width=\linewidth]{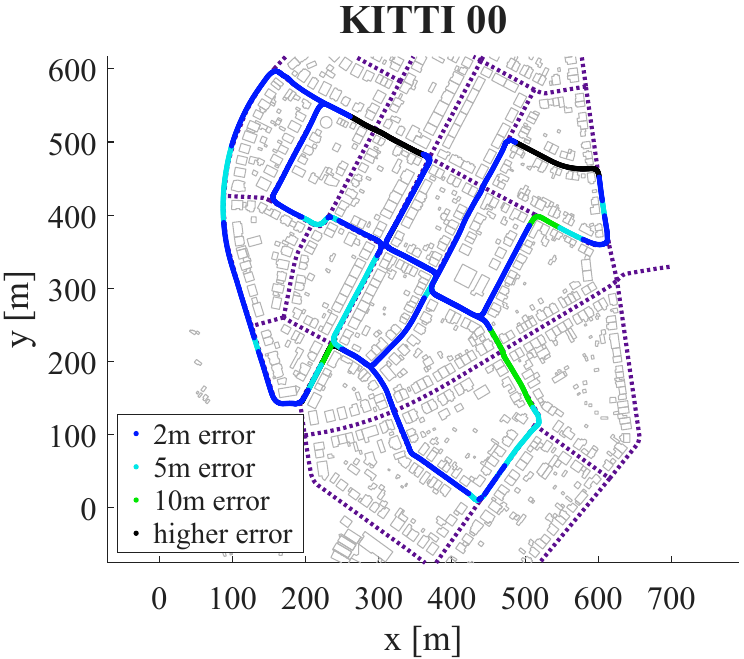}
        \label{fig:sub1}
    \end{subfigure} 
    \begin{subfigure}[t]{0.245\linewidth}
        \centering
        \includegraphics[width=\linewidth]{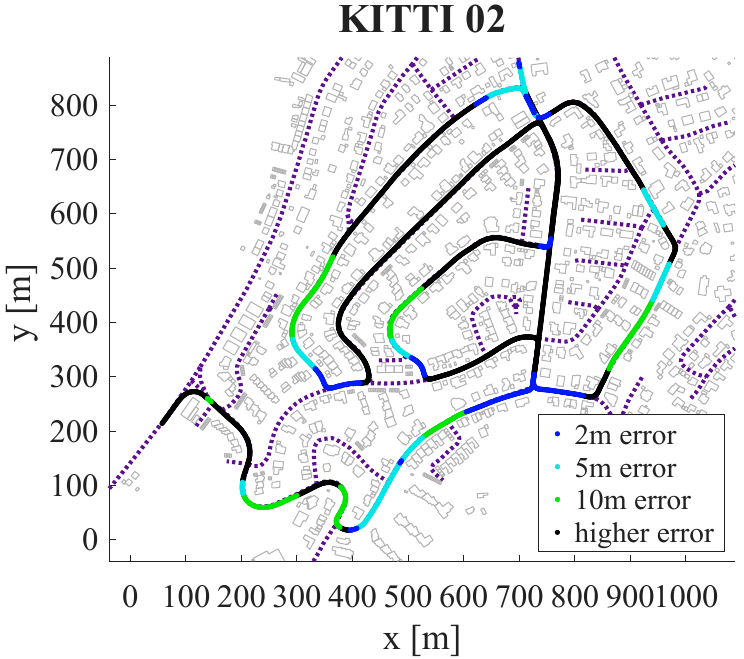}
        \label{fig:sub2}
    \end{subfigure}
    \begin{subfigure}[t]{0.245\linewidth}
        \centering
        \includegraphics[width=\linewidth]{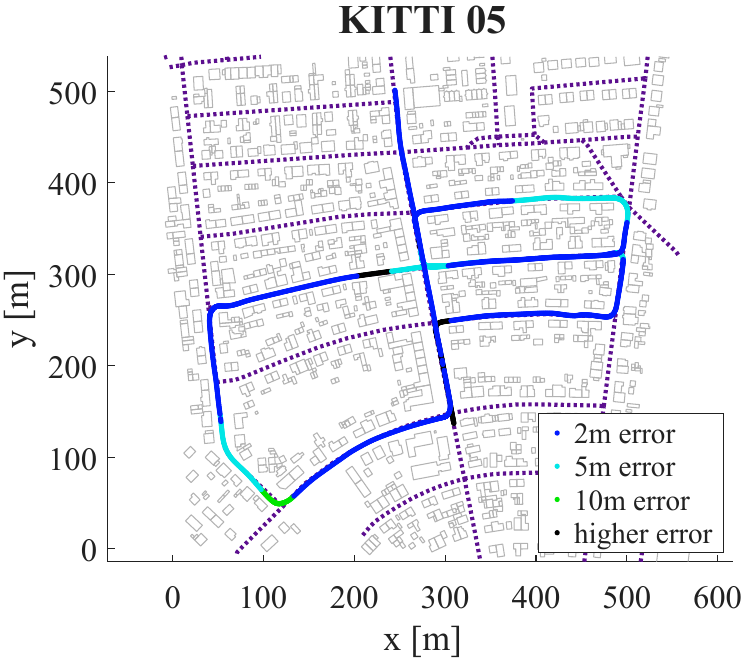}
        \label{fig:sub1}
    \end{subfigure} 
    \begin{subfigure}[t]{0.245\linewidth}
        \centering
        \includegraphics[width=\linewidth]{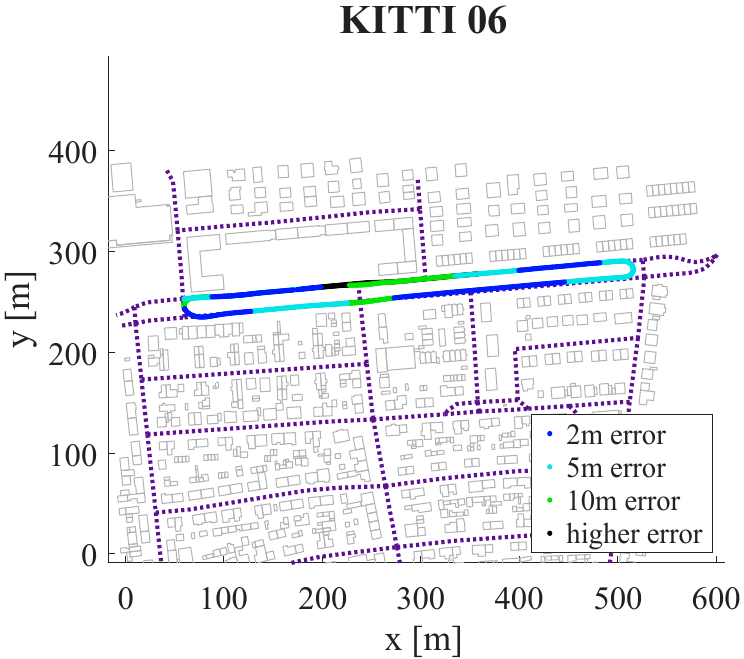}
        \label{fig:sub2}
    \end{subfigure}
    \vskip-0.6em 
    \begin{subfigure}[b]{0.245\linewidth}
        \centering
        \includegraphics[width=\linewidth]{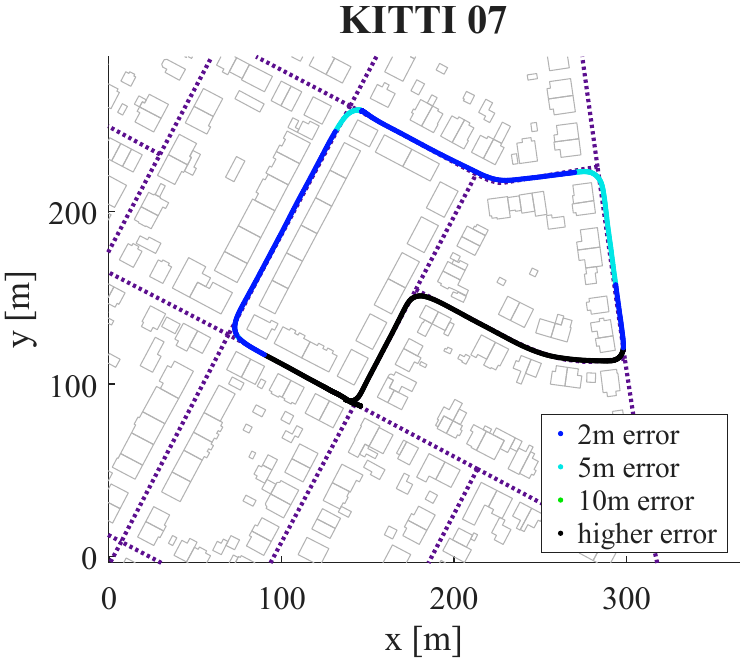}
        \label{fig:sub1}
    \end{subfigure} 
    \begin{subfigure}[b]{0.245\linewidth}
        \centering
        \includegraphics[width=\linewidth]{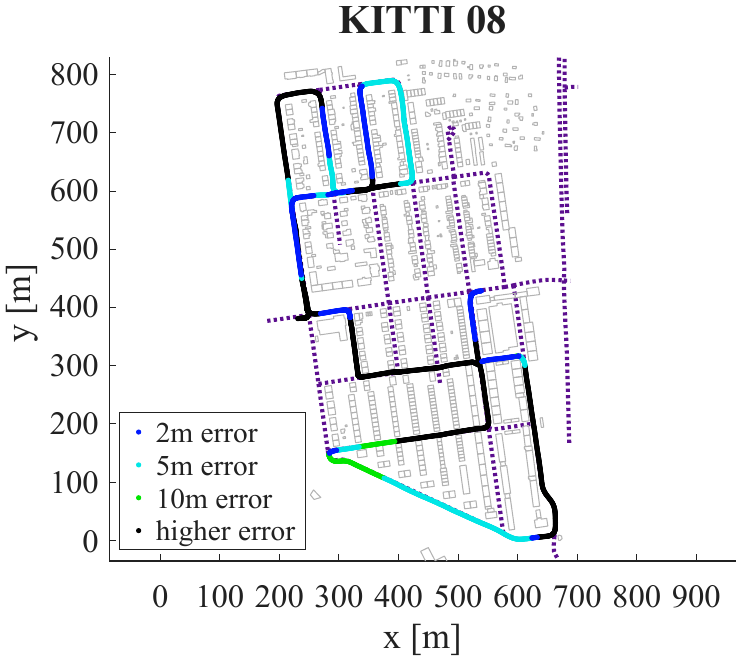}
        \label{fig:sub2}
    \end{subfigure}
    \begin{subfigure}[b]{0.245\linewidth}
        \centering
        \includegraphics[width=\linewidth]{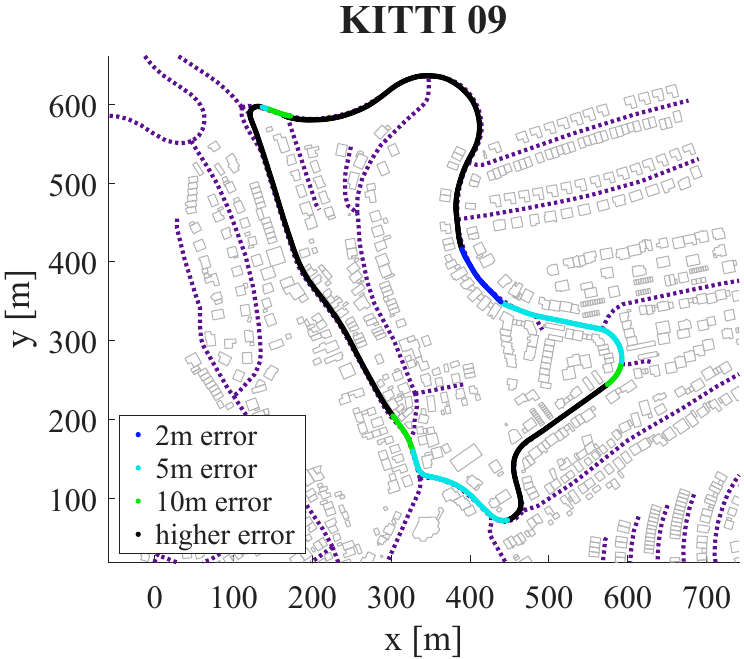}
        \label{fig:sub1}
    \end{subfigure} 
    \begin{subfigure}[b]{0.245\linewidth}
        \centering
        \includegraphics[width=\linewidth]{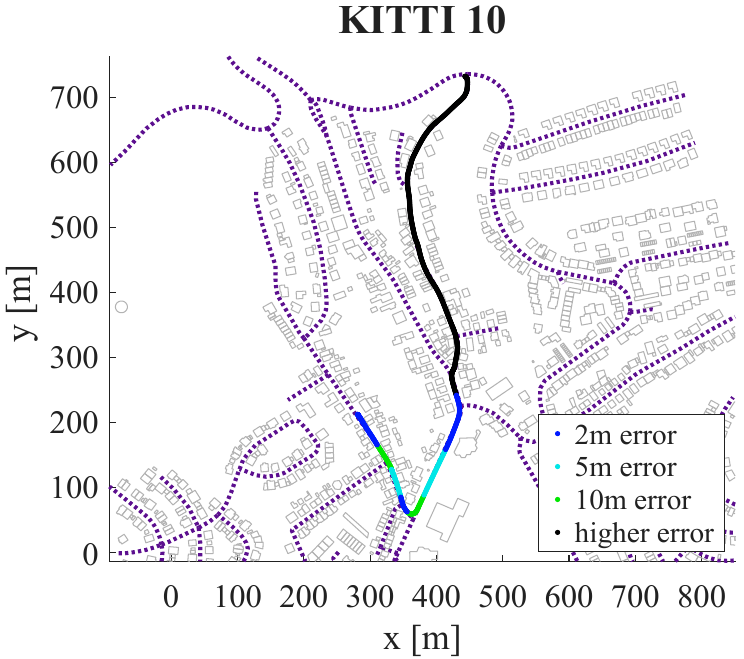}
        \label{fig:sub2}
    \end{subfigure}
    \vskip-1em 
    \caption{Vehicle global localization on KITTI dataset sequences. Blue, cyan, and green indicate frames with position errors within 2 meters, 5 meters, and 10 meters of the ground-truth position, respectively. Black indicates frames with higher errors.}
    \label{fig:gloc_8seqs}
\end{figure*}

\begin{table*}[t]
    \centering
    \caption{Recall@$5$m (\%) of global localization on KITTI sequences and the weighted average.}
        \begin{tabular}{c | c c c c c c c c | c} 
            \toprule
            Method & 00 & 02 & 05 & 06 & 07 & 08 & 09 & 10 & Average \\
            
            \midrule 
            OSM Context \cite{cho_openstreetmap-based_2022} & 48.34 & 1.85 & 37.52 & 62.13 & 37.33 & 29.89 & 22.25 & 11.66 & 29.12 \\
            BDF \cite{li2024lidar} & 35.94 & 1.16 & 21.26 & 48.95 & 40.05 & 16.68 & 9.93 & 9.08 & 19.97 \\ 
            
            \midrule
            InterKey (ours) & \textbf{81.15} & \textbf{25.83} & \textbf{89.03} & \textbf{65.03} & \textbf{54.77} & \textbf{45.69} & \textbf{29.23} & \textbf{30.39} & \textbf{54.00} \\
            \bottomrule
        \end{tabular}
    \vspace{-4mm}
    \label{tab:gloc_eval}
\end{table*}

The proposed method outperforms the baselines because it is more tailored to our context. 
To accommodate the OSM data, the top-view images in this study are represented in binary form. The BRISK keypoint descriptor, which relies on comparing gray values between specific sampled points, only performs well on images with many gray levels. In this case, the full spatial sampling principle helped the proposed method and the Scan Context approach to give better results. 
On the other hand, the top-view images in this study are generated from a series of consecutive scans, which makes the spatial distribution of points relatively uniform. The sampling pattern of Scan Context is designed for a single scan, where the point density is higher in the inner region. As a result, features in the far range receive low weight in the overall descriptor. The proposed method, with its more area-equalized sampling pattern, addresses this issue and thereby achieves better performance. 
The above experimental results have validated these designs.

\subsection{Vehicle Global Localization Performance} \label{sec:exp_eval_loc}

\subsubsection{Evaluation metrics}

We evaluate global localization accuracy using the standard Recall@$K$m metric. At each sensor frame $k$, we compute the Euclidean distance between the estimated vehicle position and the ground-truth vehicle position in the global reference system as:
\begin{equation}
    \Delta^{\mathrm{e}}_k = \vert| {}_{G}\mathbf{p}_{{L}_k} - {}_{G}\overline{\mathbf{p}}_{L_k} \vert|
\end{equation}
where ${}_{G}\overline{\mathbf{p}}_{L_k}$ is the ground-truth vehicle position given by the test dataset.
Recall@$K$m is defined as the proportion of frames where this distance is less than $K$ meters.

\subsubsection{Baselines}

We compared the proposed method with OSM Context \cite{cho_openstreetmap-based_2022} and BDF \cite{li2024lidar}, two state-of-the-art methods for point-cloud-to-OSM global localization. Other methods \cite{lee2024autonomous, kang2025opal} were excluded because they report results only on a small portion of the KITTI dataset and do not provide open-source code for re-implementation. 
To obtain comparative results, we refer to the results presented in the corresponding papers and use the same experimental setup in our approach. This includes the same KITTI data sequences and predicted semantic labels provided by RangeNet++. For OSM Context, we regard the top-1 localization accuracy as equivalent to the Recall@$5$m metric, since they used a $5$m threshold to define success. We also selected the best configuration of OSM Context with a bin length of $5$m.

\subsubsection{Results}

Fig. \ref{fig:gloc_8seqs} visualizes the proposed method's performance on the KITTI dataset under various road network and building configurations. 
Table \ref{tab:gloc_eval} provides a quantitative comparison between methods. The proposed method outperforms the baselines across all test sequences, achieving an average Recall@$5$m that is $1.85$ and $2.7$ times of that of OSM Context and BDF, respectively.

The difference in performance between the presented methods is a result of several factors. 
First, many places along the road have low measure of distinctiveness, being either featureless or highly similar to other locations in the nearby area. Consequently, place recognition at $1$-meter intervals, as in the baselines, performs poorly. Moreover, such dense global localization is unnecessary, as modern odometry methods achieve errors of less than $1$ meter per $100$ meters traveled \cite{jonnavithula2021lidar, agostinho2022survey}. Instead, we choose to match highly distinctive intersection features, which are several hundred meters apart, and project odometry to the global reference frame. This approach yields higher accuracy while reducing the cost of matching computations and database maintenance.
Another contributing factor is that trees obscure many buildings on the roadside. In these cases, relying only on building data, as in the baseline methods, leads to poor recognition. The proposed method addressed this issue by combining both road and building imprints in the descriptor construction. Sequence 02, where most of the buildings are obscured, shows the greatest improvement.











\section{Conclusion}


We introduced InterKey, a novel framework for global vehicle localization that leverages road intersections as cross-modal keypoints between point clouds and OSM. By jointly encoding road and building imprints into compact binary descriptors and introducing discrepancy mitigation, orientation determination, and area-equalized sampling strategies, our method effectively bridges the modality gap between perception data and map abstractions.
Experiments on the KITTI dataset demonstrated that InterKey achieves state-of-the-art performance, significantly outperforming existing OSM-based localization methods such as OSM Context and BDF. These results highlight the advantages of focusing on distinctive and robust landmarks, such as intersections, rather than relying solely on building features or dense global descriptors.
Looking ahead, we plan to extend InterKey by incorporating visual cues and multimodal data to improve descriptor robustness, and by exploring purely camera- or radar-based implementations. Such extensions will broaden its applicability across diverse sensing platforms and further enhance the scalability of OSM-based global localization.






\section*{ACKNOWLEDGMENT}

The authors would like to thank Dr. Thien Hoang Nguyen for the insightful discussions and valuable feedback.


\bibliographystyle{ieeetr} 
\bibliography{references/selected} 




\end{document}